\documentclass[10pt,twocolumn,letterpaper]{article}

\usepackage{cvpr}
\usepackage{times}
\usepackage{epsfig}
\usepackage{graphicx}
\usepackage{amsmath}
\usepackage{amssymb}
\usepackage{caption}
\usepackage{url}
\usepackage[breaklinks=true,bookmarks=false]{hyperref}

\cvprfinalcopy 

\def\cvprPaperID{108} 

\ifcvprfinal\fi

\begin{document}

\title{WiCV@CVPR2023: The Eleventh Women In Computer Vision Workshop at the Annual CVPR Conference}

\author{ Doris Antensteiner$^1$, Marah Halawa$^2$, Asra Aslam$^3$, Ivaxi Sheth$^4$,\\ Sachini Herath$^5$, Ziqi Huang$^6$, Sunnie S. Y. Kim$^7$, Aparna Akula$^8$, Xin Wang$^9$ \\\\ $^1$Austrian Institute of Technology, $^2$Technical University of Berlin, $^3$University of Leeds,\\$^4$CISPA – Helmholtz Center for Information Security, $^5$Simon Fraser University,  \\ $^6$S-Lab, Nanyang Technological University, $^7$Princeton University, \\ $^8$CSIR – Central Scientific Instruments Organisation, $^9$Microsoft\\
 \tt\small wicvcvpr2023-organizers@googlegroups.com
}
\maketitle

\begin{abstract}
\thispagestyle{empty}
In this paper, we present the details of Women in Computer Vision Workshop - WiCV 2023, organized alongside the hybrid CVPR 2023 in Vancouver, Canada.  WiCV aims to amplify the voices of underrepresented women in the computer vision community, fostering increased visibility in both academia and industry. We believe that such events play a vital role in addressing gender imbalances within the field. The annual WiCV@CVPR workshop offers a)~opportunity for collaboration between researchers from minority groups, b)~mentorship for female junior researchers,  c)~financial support to presenters to alleviate finanacial burdens and d)~a diverse array of role models who can inspire younger researchers at the outset of their careers. In this paper, we present a comprehensive report on the workshop program, historical trends from the past WiCV@CVPR events, and a summary of statistics related to presenters, attendees, and sponsorship for the WiCV 2023 workshop.

\end{abstract}

\section{Introduction}
Despite remarkable progress in various computer vision research areas in recent years, the field still grapples with a persistent lack of diversity and inclusion. While the field of computer vision rapidly expands, female researchers remain underrepresented in the area, constituting only a small amount of professionals in both academia and industry. Due to this, many female computer vision researchers can feel isolated in workspaces which remain unbalanced due to the lack of inclusion.

The WiCV workshop is a gathering designed for all individuals, irrespective of gender, engaged in computer vision research. It aims to appeal to researchers at all levels, including established researchers in both industry and academia (e.g. faculty or postdocs), graduate students pursuing a Masters or PhD, as well as undergraduates interested in research.  The overarching goal is to enhance the visibility and recognition of female computer vision researchers across these diverse career stages, reaching women from various backgrounds in educational and industrial settings worldwide.

There are three key objectives of the WiCV workshop:

\paragraph{Networking and Mentoring}

The first objective is to expand the WiCV network and facilitate interactions between members of this network. This includes female students learning from seasoned professionals who share career advice and experiences. A mentoring banquet held alongside the workshop provides a casual environment for junior and senior women in computer vision to meet, exchange ideas and form mentoring or research relationships.

\paragraph{Raising Visibility}

The workshop's second objective is to elevate the visibility of women in computer vision, both at junior and senior levels. Senior researchers are invited to give high quality keynote talks on their research, while junior researchers are encouraged to submit their recent or ongoing work, with many of these being selected for oral or poster presentation through a rigorous peer review process. This empowers junior female researchers to gain experience presenting their work in a professional yet supportive setting. The workshop aims for diversity not only in research topics but also in the backgrounds of presenters. Additionally, a panel discussion provides a platform for female colleagues to address topics of inclusion and diversity.

\paragraph{Supporting Junior Researchers}

Finally, the third objective is to offer junior female researchers the opportunity to attend a major computer vision conference that might otherwise be financially inaccessible. This is made possible through travel grants awarded to junior researchers who present their work during the workshop's poster session. These grants not only enable participation in the WiCV workshop but also provide access to the broader CVPR conference.

\section{Workshop Program}
\label{program}
The workshop program featured a diverse array of sessions, including 4 keynotes, 6 oral presentations, 34 poster presentations, a panel discussion, and a mentoring session. Consistent with previous years, our keynote speakers were carefully selected to ensure diversity in terms of the topics that were covered, their backgrounds, whether they work in academia or industry, and their seniority.  This deliberate choice of diverse speakers is of paramount importance, as it offers junior researchers a multitude of potential role models with whom they can resonate and, in turn, envision their unique career paths.

The workshop schedule at CVPR 2023 featured a diverse range of sessions and activities, including:
\begin{itemize}
    \item Introduction
    
    \item Invited Talk 1: Angel Chang (Simon Fraser University and Canada CIFAR AI), \textit{Connecting 3D and Language}
    
    \item Oral Session 1
    \begin{itemize}
        \item Laia Tarrés, \textit{Sign Language Translation for Instructional Videos}
        \item Meghna Kapoor, \textit{Underwater Moving Object Detection using an End-to-End Encoder-Decoder Architecture and GraphSage with Aggregator and Refactoring}
    \end{itemize}
    
    \item Invited Talk 2: Devi Parikh (Generative AI Lead at Meta), \textit{Multimodal Generative AI (AI for creativity)}
    
    \item Sponsors Exhibition (in person)
    
    \item Oral Session 2
    \begin{itemize}
        \item Deblina Bhattacharjee, \textit{Dense Multitask Learning to Reconfigure Comics}
        \item Maryam Daniali, \textit{Perception Over Time: Temporal Dynamics for Robust Image Understanding}
    \end{itemize}
    
    \item Invited Talk 3: Judy Hoffman (School of Interactive Computing at Georgia Tech), \textit{Efficient and Reliable Vision Models}
    
    \item Sponsors Exhibition (in person)
    
    \item Oral Session 3
    \begin{itemize}
        \item Zoya Shafique, \textit{Nonverbal Communication Cue Recognition: A Pathway to More Accessible Communication}
        \item Sudha Velusamy, \textit{A Light-Weight Human Eye Fixation Solution for Smartphone Applications}
    \end{itemize}
    
    \item Invited Talk 4: Kristen Grauman (University of Texas at Austin and Facebook AI Research (FAIR)), \textit{Human-object interactions in first person video}
    
    \item Panel Discussion by Abby Stylianou, Angel Chang, Devi Parikh, Ilke Demir, Judy Hoffman, and Kristen Grauman
    
    \item Poster Session (in person as well as virtual)
    
    \item Closing Remarks
    
    \item Mentoring Session, Talks, and Dinner (in person)
    \begin{itemize}       
        \item Invited talk:
        \begin{enumerate}
            \item Abby Stylianou (Saint Louis University and Taylor Geospatial Institute) on \textit{A bit about my research and finding your own path to happiness and success in this field even if it looks different}
            \item Ilke Demir (Intel Corporation) on the intriguing topic \textit{Be you}
        \end{enumerate}

         \item In-Person mentoring session and dinner: Akshita Gupta (University of Guelph), Dima Damen (University of Bristol), Achal Dave (Toyota Research Institute), Federica Arrigoni (Politecnico di Milano), Hilde Kuehne (University on Bonn), Nikhila Ravi (Meta), Katherine Liu (Toyota Research Institute), Lilly Ball (Apple), Dian Chen (Toyota Research Institute), Shuyang Cheng (Cruise LLC), Jeremy Jones (Apple), Zuzana Kukelova (Czech Technical University in Prague). 
         
        \item Virtual mentoring session on Zoom: Yifan Liu (ETH Zürich), and Dena Bazazian (University of Plymouth).
    \end{itemize}
\end{itemize}

\subsection{Hybrid Setting}
This year, our organizational approach underwent slight adjustments due to CVPR 2023 being held in a hybrid setting, accommodating both in-person and virtual attendance. We took deliberate steps to sure that the virtual WiCV workshop was an engaging and interactive event. To achieve this, we took the following steps: Talks, oral sessions, and the panel were streamed via Zoom for virtual attendances. The poster session was repeated virtually a week after the conference, mirroring the format of the main conference. We also facilitated online mentoring sessions via Zoom, catering to mentors and mentees who could only participate virtually. 

\section{Workshop Statistics}

The first edition of the Women in Computer Vision (WiCV) workshop was held in conjunction with CVPR 2015. Over the years, both the participation rate and the quality of submissions to WiCV have steadily increased.  
Following the examples from the editions held in previous years \cite{asra2023wicv, antensteiner2022wicv, doughty2021wicv, Amerini19, Akata18, Demir18} we have continued to curate top-quality submissions into our workshop proceedings. By providing oral and poster presenters the opportunity to publish their work in the conference's proceedings, we aim to further boost the visibility of female researchers. 

This year, the workshop was held as a half-day in-person event with hybrid options, while the virtual component was hosted via Topia and Zoom. The in-person gathering took place at the Vancouver Convention Centre in Canada. Senior and junior researchers were invited to present their work, including the poster presentations detailed in the previous Section \ref{program}.\\

The organizers for this year's WiCV workshop come from diverse backgrounds in both academia and industry, representing various institutions across different time zones. Their diverse backgrounds and wide-ranging research areas have enriched the organizing committee's perspectives and contributed to a well-rounded approach. Their broad range of research interests in computer vision and machine learning encompass video understanding, object detection, non-verbal communication, open-source benchmark datasets, activity recognition, anomaly detection, autoencoders, generalization, captioning, 3D Point Cloud, medical imaging and vision for robotics.

\begin{figure}[h]
\centering
\includegraphics[width=1\linewidth]{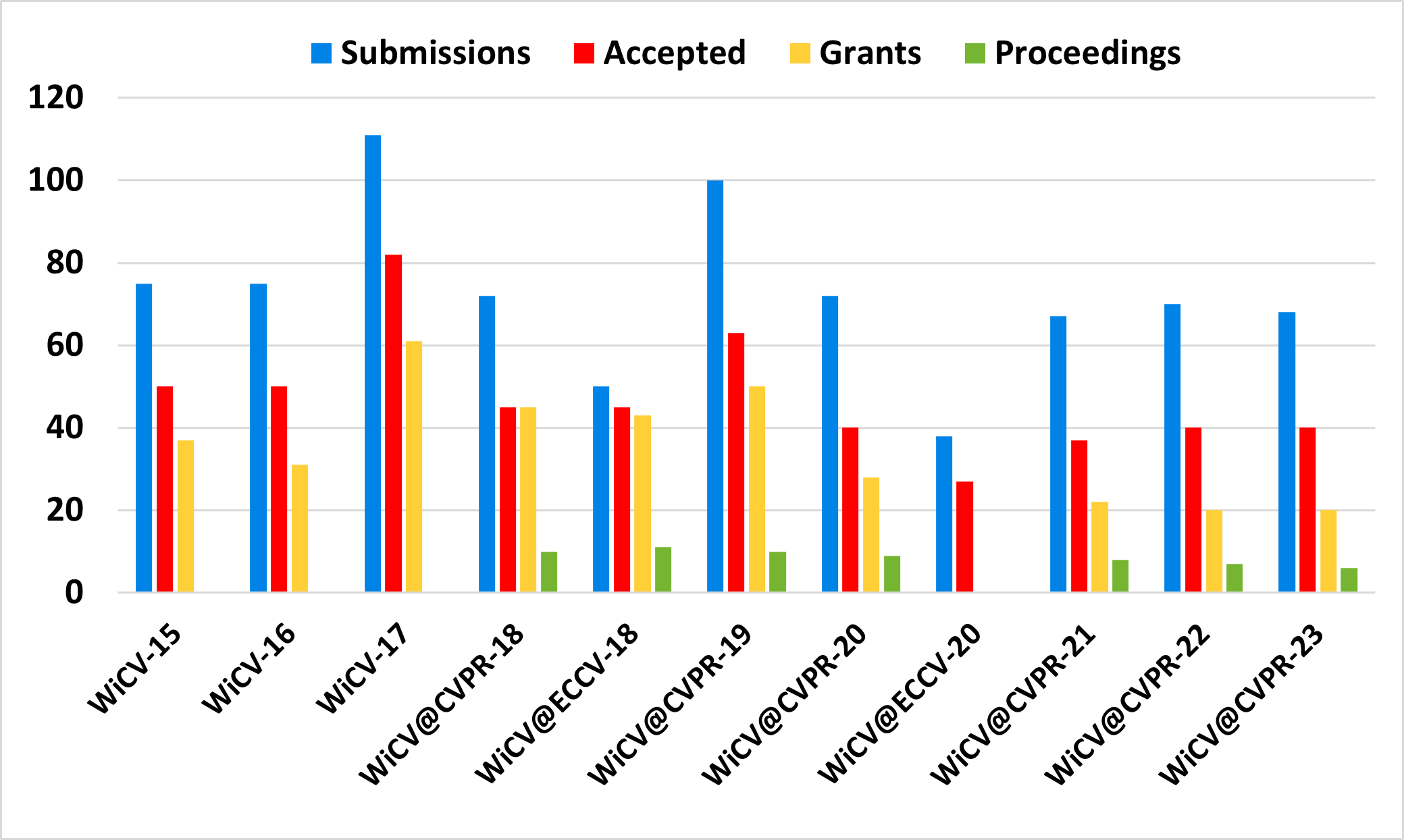}
\captionof{figure}{\textbf{WiCV Submissions.} The number of submissions over the past years of WiCV.}
\label{fig:sub}
\end{figure}

This year, we received 68 high-quality submissions from a wide range of topics and institutions, which is on par with WiCV@CVPR22. The most popular topics included deep learning architectures and techniques followed by video action and event recognition, segmentation and shape analysis, and medical applications. 
Out of the 68 submissions, 61 underwent the review process. Six papers were selected as oral presentations and inclusion in the CVPR23 workshop proceedings, while 34 papers were chosen for poster presentations. The comparison with previous years is presented in Figure~\ref{fig:sub}. Thanks to the diligent efforts of an interdisciplinary program committee comprising 41 reviewers, the submitted papers received thorough evaluations and valuable feedback. Additionally, during the mentoring session, 40 mentees received in-person guidance from 8 mentors, and 5 mentees attended virtual sessions with 2 mentors in separate meetings via Zoom.

This year, we continued the WiCV tradition from previous workshops \cite{Akata18,Amerini19,Demir18,doughty2021wicv, goel2022wicv} by providing grants to assist the authors of accepted submissions in participating in the workshop. These grants covered a range of expenses, with the specifics varying for each attendee depending on their individual needs, including, for example, conference registration fees, round-trip flight itineraries, and two days of accommodation for all authors of accepted submissions who requested funding.

\begin{figure}
\centering
\includegraphics[width=1\linewidth]{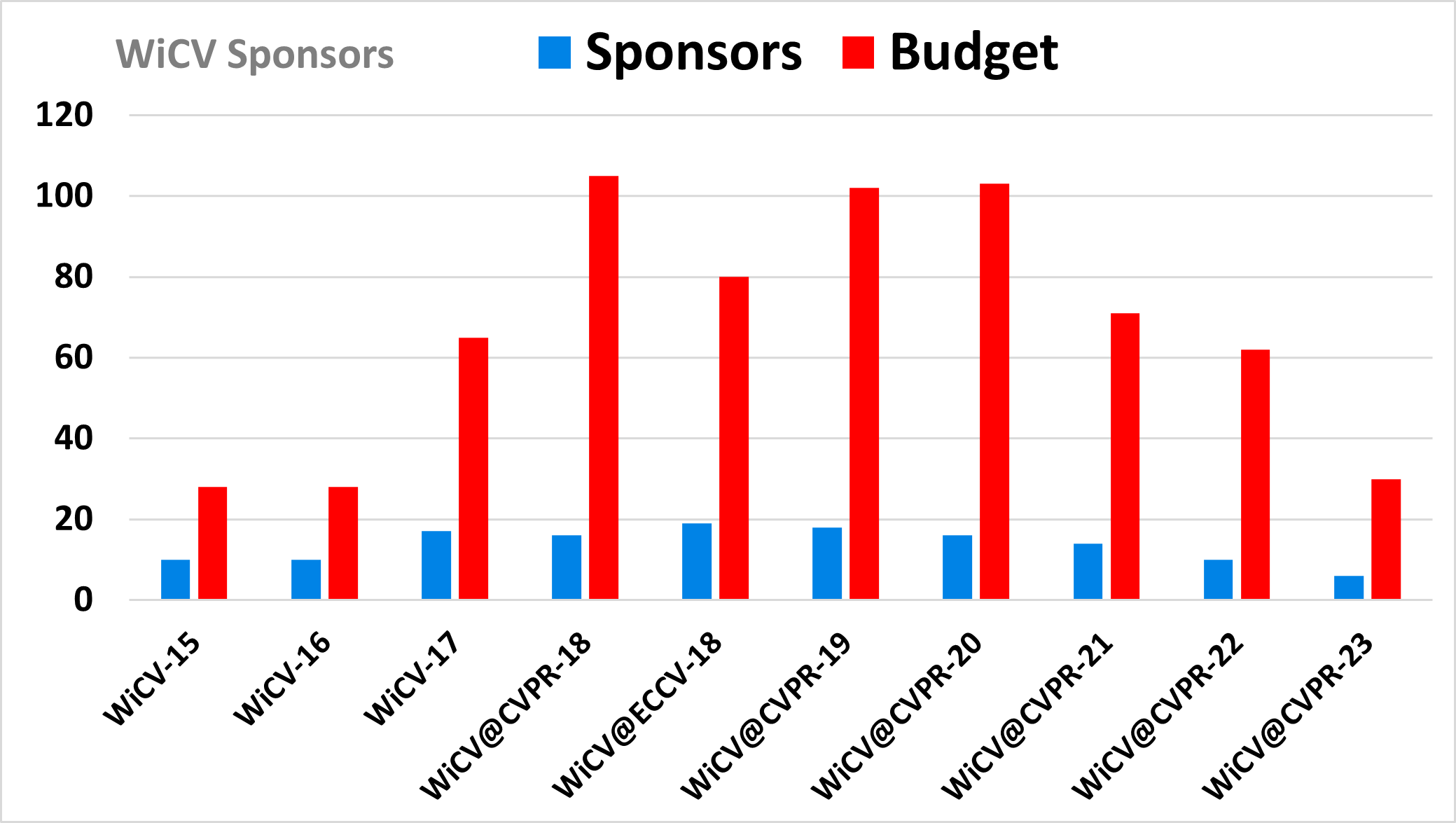}
\captionof{figure}{\textbf{WiCV Sponsors.} The number of sponsors and the amount of sponsorship for WiCV. The amount is expressed in US dollar (USD).}
\label{fig:spo}
\end{figure}

The total sponsorship for this year's workshop amounted to \$30,000 USD, with contributions of 6 sponsors, meeting our target. In Figure~\ref{fig:spo} you can find the details with respect to the past years. 

\section{Conclusions}
WiCV at CVPR 2023 has once again proven to be a valuable opportunity for presenters, participants, and organizers, providing a platform to unite the community. It continues to address the persistent issue of gender balance in our field, and we believe it has played a significant role in strengthening the community. It provided an opportunity for people to connect from all over the world from the comfort of their personal spaces. With a high number of paper submissions and even greater number of attendees, we anticipate that the workshop will continue the positive trajectory of previous years, fostering a stronger sense of community, increased visibility, and inclusive support and encouragement for all female researchers in academia and industry. Moreover, WiCV Members participated in the Diversity \& Inclusion Social event at CVPR. Furthermore, WiCV also got featured by in the CVPR~2023 Magazine \cite{cvprmagazine}.

\section{Acknowledgments}
We express our sincere gratitude to our sponsors, including our Platinum sponsors: Toyota Research Institute and Apple, as well as our Silver Sponsor: Cruise, and Bronze sponsors: Deepmind, Meshcapade, and Snap Inc. Our appreciation also extends to the San Francisco Study Center, our fiscal sponsor, for their invaluable assistance in managing sponsorships and travel awards. We are thankful for the support and knowledge-sharing from organizers of previous WiCV workshops, without whom this WiCV event would not have been possible. Finally, we extend our heartfelt thanks to the dedicated program committee, authors, reviewers, submitters, and all participants for their valuable contributions to the WiCV network community.

\section{Contact}
\noindent \textbf{Website}: \url{https://sites.google.com/view/wicvcvpr2023/home}\\
\textbf{E-mail}: wicvcvpr2023-organizers@googlegroups.com\\
\textbf{Facebook}: \url{https://www.facebook.com/WomenInComputerVision/}\\
\textbf{Twitter}: \url{https://twitter.com/wicvworkshop}\\
\textbf{Google group}: women-in-computer-vision@googlegroups.com\\

{\small
\bibliographystyle{ieee}
\bibliography{egbib}
}

\end{document}